\documentclass[runningheads]{llncs}

\usepackage[utf8]{inputenc}
\usepackage{booktabs}
\usepackage{marvosym}
\usepackage{enumitem}
\usepackage{multirow}
\usepackage{siunitx}
\usepackage{graphicx}
\usepackage{amssymb}
\usepackage{amsmath}
\usepackage{booktabs}
\usepackage[T1]{fontenc}
 \usepackage{hyperref}
\usepackage{graphicx,verbatim}
\usepackage[nolist,nohyperlinks]{acronym}
\begin{document}

\title{Track2Map: Online Deformable SLAM with Motion-Aware Pose Optimization in Robotic Surgery}
\titlerunning{Track2Map}
%

\author{
Tianyi Song\inst{1}$^\star$\textsuperscript{(\Letter)} \and
Sierra Bonilla\inst{1}$^\star$ \and
Xinwei Ju\inst{1} \and 
Evangelos Mazomenos\inst{1} \and 
Danail Stoyanov\inst{1} \and 
Adam Schmidt\inst{2} \and
Omid Mohareri\inst{2} \and
Sophia Bano\inst{1} \and 
Francisco Vasconcelos\inst{1}
}

\authorrunning{T. Song et al.}

\institute{
Department of Computer Science and UCL Hawkes Institute, University College London, London WC1E 6BT, UK
\and
Intuitive Surgical, Inc., Sunnyvale, CA 94086, USA 
\email{\{tianyi.song.24,sierra.bonilla.21\}@ucl.ac.uk}
}

\acrodef{2d}[2D]{two-dimensional}
\acrodef{3d}[3D]{three-dimensional}
\acrodef{sfm}[SfM]{structure-from-motion}
\acrodef{sift}[SIFT]{Scale-Invariant Feature Transform}
\acrodef{ransac}[RANSAC]{random sample consensus}
\acrodef{lift}[LIFT]{Learned Invariant Feature Transform}
\acrodef{cnn}[CNN]{convolutional neural network}
\acrodef{se3}[SE(3)]{Special Euclidean group in 3D}
\acrodef{RAMIS}[RAMIS]{robot-assisted minimally invasive surgery}
\acrodef{mft}[MFT]{Multi-Flow dense Tracker}
\acrodef{raft}[RAFT]{Recurrent All-Pairs Field Transforms}
\acrodef{sota}[SOTA]{state-of-the-art}
\acrodef{slam}[SLAM]{Simultaneous Localisation and Mapping}

\maketitle

\begingroup
\renewcommand\thefootnote{$\star$}
\footnotetext{These authors contributed equally.}
\endgroup

\begin{abstract}
Gaussian splatting is the current state-of-the-art for dense, deformable 3D anatomy reconstruction in \ac{RAMIS}; however, most pipelines are offline and depend on accurate camera trajectory priors (often from robotic kinematics), limiting applicability when priors are missing or noisy. To address these limitations, we propose \textbf{Track2Map}, an online 3D Gaussian Splatting pipeline that jointly optimizes camera trajectory and 3D deformable scene representation directly from surgical video. Track2Map is therefore capable of robust 3D reconstructions when camera trajectory priors are either absent or noisy, and due to its online nature it effectively works as a \ac{slam} method. To stabilize optimization in the presence of tissue motion and ambiguous visual cues, we introduce a track-anchored deformation initialization using dense 2D point tracks. Track statistics are further utilized to disentangle camera motion from scene deformation by detecting static camera periods and reducing drift during incremental mapping. Experiments on StereoMIS show improved reconstruction quality and camera trajectory against competing \ac{slam} methods, as well as compared to non-SLAM methods that utilize camera trajectory priors. The code is available at \href{https://track2map.github.io/}{https://track2map.github.io/}.
\keywords{Robot-assisted surgery \and Stereo endoscopy \and Online 3D reconstruction \and Camera pose optimization}
\end{abstract}

\section{Introduction}

\begin{figure*}[t]
  \centering
  \includegraphics[width=\textwidth]{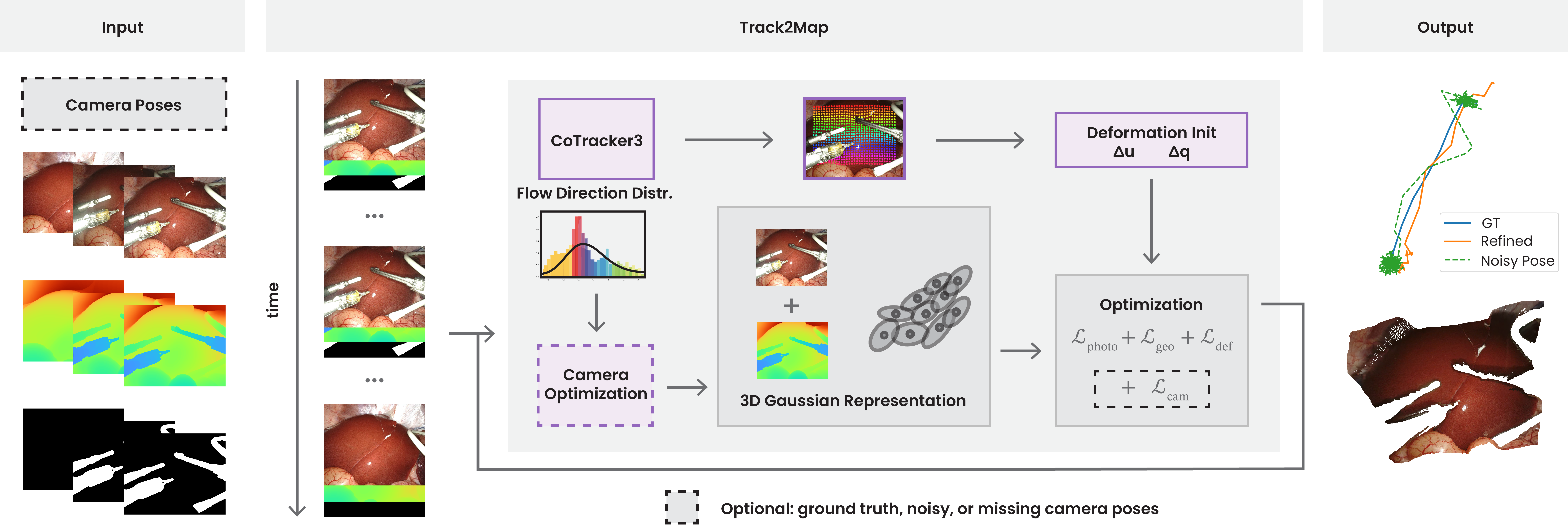}
  \caption{Given stereo RGB frames, depth maps, and \textit{optional} (clean/noisy) pose extrinsics/intrinsics and tool masks, Track2Map jointly performs 2D tracking and online deformable 3D reconstruction.~\cite{karaev2025cotracker3} provides 2D correspondences that are lifted to 3D and used to estimate/refine camera motion via motion-gated pose optimization and initialize sparse anchor-based deformation. A Gaussian representation like~\cite{hayoz2024online} is updated online. The system outputs refined poses and metric reconstruction; 2D/3D tracks are available as by-products of the correspondence module and 3D lifting.}
  \label{fig:method_first_pass}
\end{figure*}

Online 3D reconstruction of deformable tissue is a key prerequisite for developing a range of computer guidance techniques in \ac{RAMIS}, including characterisation of instrument-tissue interactions, autonomous manipulation, and stable augmented reality overlays. However, achieving metric, stable reconstruction in \ac{RAMIS} remains difficult because the scene is non-rigid, frequently occluded by instruments, and captured with limited viewpoint variation. 

3D reconstruction methods have progressed from classical SfM~\cite{schonberger2016structure} to neural radiance fields~\cite{mildenhall2021nerf} and 3D Gaussian Splatting (3DGS)~\cite{kerbl20233d} representations tailored to endoscopy~\cite{bonilla2024gaussian,wang2024endogslam,diff2dgs}. However, extending these representations to deformable surgical scenes introduces a fundamental coupling problem: camera pose, non-rigid deformation, and appearance/geometry estimation must be estimated jointly, but supervision and priors are often unreliable. Most surgical neural/3DGS methods either assume static or known camera motion, or depend on relatively clean pose inputs to prevent drift in deformable sequences \cite{hayoz2024online,huang2024endo}.
In \ac{RAMIS}, camera pose is often calculated from robot joint kinematics, which can be inaccurate due to calibration errors and measurement noise, and may also not always be available in a clinical setting. Therefore, camera motion availability and quality varies widely across available surgical datasets; some provide accurate trajectories~\cite{hayoz2023learning}, others provide noisy priors~\cite{bobrow2023colonoscopy}, and many tracking benchmarks provide none ~\cite{cartucho2024surgt,schmidt2024tracking}. This motivates deformable reconstruction systems that can exploit pose priors when available, remain stable when priors are corrupted, and can initialize and operate when pose is absent.

Joint estimation of camera motion and deformable structure has a long history in NRSfM and non-rigid SLAM, including locally NRSfM~\cite{taylor2010non}, sequential physical-prior NRSfM~\cite{agudo2015sequential}, online self-calibrating NRSfM~\cite{agudo2021total}, and deformable SLAM~\cite{xu2022active}. These works establish the broader paradigm of coupling camera motion with non-rigid scene structure, while our setting targets online dense 3DGS reconstruction from surgical stereo video with optional or noisy pose priors. In surgical neural/3DGS mapping, recent systems have begun to jointly optimize camera motion and deformable scene structure~\cite{wang2025endosd,wu2025endoflow}. However, a key failure mode persists in RAMIS: endoscope motion is often intermittent, so long segments have a nearly static camera while tools deform tissue. In these intervals, image motion is dominated by local deformation and instrument motion, and indiscriminate pose optimization can incorrectly attribute tissue dynamics to camera motion, causing trajectory drift and destabilizing reconstruction. Our work addresses these issues by introducing two key additions.
%
%
First, rather than completely eliminating pose priors, we use them as initial estimates whenever available, making our system flexible enough to operate both as a \ac{slam} system and as an online 3D reconstruction pipeline that can handle noisy camera priors. Second, we use dense 2D point tracking to disentangle camera motion from tissue deformation. This allows us to detect when the camera view is static and dynamically introduce camera pose constraints that improve 3D reconstruction quality and reduce camera trajectory drift. Modern methods for dense 2D point tracking build on deep optical flow~\cite{teed2020raft} and sequence models~\cite{neoral2024mft,karaev2025cotracker3}. Surgical extensions improve robustness and efficiency~\cite{karaoglu2025litetracker,seenivasan2025endotracker,schmidt2024tracking}; in this work, we use CoTracker3~\cite{karaev2025cotracker3} as our 2D tracking initialization for Gaussian Splatting after empirically verifying its state-of-the-art performance on the STIR2024 benchmark \cite{schmidt2024tracking}.

Experiments on StereoMIS~\cite{hayoz2023learning} demonstrate Track2Map's improved camera tracking and scene reconstruction over baselines including~\cite{neoral2024mft,karaev2025cotracker3}, and robustness under pose corruption and absence. Our core contributions are as follows:
\begin{itemize}[leftmargin=*, topsep=0pt, itemsep=2pt]

\item An online stereo 3D Gaussian reconstruction pipeline in which dense 2D correspondences are lifted to 3D anchors to initialize and constrain deformation.
\item A pose update strategy that prevents drift during camera-static periods and enables joint optimization of pose and deformation when motion is detected.
\item A unified system that operates with \emph{clean}, \emph{noisy}, or \emph{absent} pose priors without retuning, including a vision-only initialization via robust 3D alignment of tracked anchor points when priors are unavailable.
\item Improved reconstruction quality and robust pose recovery compared to prior deformable 3DGS pipelines.
\end{itemize}

\section{Method}
\label{sec:method}

Fig.~\ref{fig:method_first_pass} summarizes Track2Map's online inference loop. The central technical challenge is disentangling camera motion from non-rigid tissue motion under limited viewpoint change: inverse rendering on essentially one viewpoint (two short-baseline stereo images) provides weak constraints on pose when the scene itself also deforms, and unconstrained pose updates can easily absorb tissue deformation or tool motion, leading to significant errors. Track2Map therefore injects two procedure-driven priors into the optimization: (i) stereo depth provides metric 3D support for lifting correspondences, and (ii) endoscope motion is typically intermittent. Within \ac{RAMIS}, the endoscope arm is often stationary while surgical tools operate on the scene. Therefore, pose should be updated conservatively and only when the observed motion is consistent with camera motion rather than local deformation or tool interaction.

Following~\cite{hayoz2024online}, we represent the scene as 3D Gaussians~\cite{kerbl20233d} coupled with a sparse set of deformation control points. We use the tracked correspondences from~\cite{karaev2025cotracker3} as the primary coupling signal between tracking and mapping in three stages: (1) compute a robust per-frame motion score from track displacements and use this score to gate pose updates to occur only when global motion is detected (Sec.~\ref{sec:pose_init_update}); (2) lift the same correspondences to 3D and use their displacements to initialize a sparse anchor deformation prior, interpolated to the full set of Gaussians (Sec.~\ref{sec:anchor_deform_init}); and (3) jointly refine scene parameters (and pose when enabled) using photometric and geometric rendering losses, deformation regularization, and 2D reprojection consistency of anchors (Sec.~\ref{sec:joint_opt}).

\subsection{Motion-Aware Camera Update}
\label{sec:pose_init_update}

At frame $t$, the tracker~\cite{karaev2025cotracker3} works on a sliding window and provides 2D correspondences $\{(\mathbf{q}_k^{t-1},\,\mathbf{q}_k^t)\}_{k\in\mathcal{K}_t}$ with flow $\mathbf{f}_k^t = \mathbf{q}_k^t - \mathbf{q}_k^{t-1}$. We use these tracks to decide when pose refinement is warranted. When the camera moves, tracked points exhibit a coherent global displacement field; when the camera is stationary, observed motion is dominated by heterogeneous tissue deformation and tool interaction. So we gate pose refinement using a simple statistic of the optical-flow direction distribution. We compute the flow direction, $
\theta_k^t = \mathrm{atan2}(f_{k,y}^t,\, f_{k,x}^t).$
Empirically, camera motion produces a narrow, unimodal distribution of $\{\theta_k^t\}$ (low dispersion), whereas camera-static periods produce a broad distribution (high dispersion) due to local tissue deformation, tool motion, and physiological effects (distribution visualizations are provided in the supplementary material).
We quantify dispersion using the (circular) standard deviation of flow directions. Let
\begin{equation}
\bar{c}_t = \frac{1}{|\mathcal{K}_t|}\sum_{k\in\mathcal{K}_t}\cos\theta_k^t,
\qquad
\bar{s}_t = \frac{1}{|\mathcal{K}_t|}\sum_{k\in\mathcal{K}_t}\sin\theta_k^t,
\qquad
R_t = \sqrt{\bar{c}_t^2+\bar{s}_t^2}
\end{equation}
We then compute the circular standard deviation $
\sigma_{\theta,t} = \sqrt{-2\ln(R_t)}$. Low $\sigma_{\theta,t}$ indicates camera-dominant motion, while high $\sigma_{\theta,t}$ indicates scene motion. The motion gate is $
g_t = \mathbb{I}\!\left[\sigma_{\theta,t} \le \tau\right]$ where $\tau$ is set empirically.

When no pose prior is available, we initialize the relative motion $\Delta\mathbf{T}_t$ by aligning 3D anchor positions across frames where anchors are initialized once by sparsely subsampling gaussian centers similarly to~\cite{hayoz2024online}. 
Each tracked 2D point $\mathbf{q}_k^t$ is lifted to 3D using stereo depth and intrinsics, yielding $\mathbf{P}_k^{t-1}$ and $\mathbf{P}_k^{t}$.
We estimate $\Delta\mathbf{T}_t\in SE(3)$ via robust weighted alignment:
\begin{equation}
\Delta\mathbf{T}_t =
\arg\min_{\Delta\mathbf{T}\in SE(3)}
\sum_{k\in\mathcal{K}_t}
\omega_k\,
\rho\!\left(\bigl\|\mathbf{P}_k^{t} - \Delta\mathbf{T}\,\mathbf{P}_k^{t-1}\bigr\|_2^2\right)
\label{eq:delta_t}
\end{equation}
where $\omega_k$ is the tracker confidence and $\rho(\cdot)$ is a robust kernel. We set $\mathbf{T}_t^{\mathrm{init}} = \Delta\mathbf{T}_t\,\mathbf{T}_{t-1}$. If $g_t{=}0$, we freeze the pose to avoid absorbing non-rigid motion into the camera trajectory. If $g_t{=}1$, we refine pose with a small 6D increment
$\delta\boldsymbol{\xi}\in\mathbb{R}^6$ (3D translation + 3D rotation in $\mathfrak{se}(3)$),
optimized with the scene (Sec.~\ref{sec:joint_opt}) where $\exp(\cdot^{\wedge})$ projects the vector on to a valid transformation matrix:
\begin{equation}
\mathbf{T}_t =
\exp\!\left(\delta\boldsymbol{\xi}^{\wedge}\right)\,
\mathbf{T}_t^{\mathrm{init}}
\end{equation}

\subsection{Tracker-Guided Deformation Initialization}
\label{sec:anchor_deform_init}
We represent tissue deformation using the same sparse anchor set
$\mathcal{K}$ used for pose estimation (Eq.~\ref{eq:delta_t}).
Unlike prior work~\cite{hayoz2024online}, which treats control-point offsets as free variables optimized per frame, we explicitly drive the deformation field using tracked anchor motion. For each anchor $k\in\mathcal{K}$, its 3D displacement between consecutive frames is:
\begin{equation}
    \Delta\mathbf{P}_k = \mathbf{P}_k^{t} - \mathbf{P}_k^{t-1},
\end{equation}
where $\mathbf{P}_k^t$ is obtained by lifting the tracked 2D
correspondence $\mathbf{q}_k^t$ using stereo depth and the current
camera pose $\mathbf{T}_t$.
The initial deformation offset $\Delta\mathbf{x}_i^{(0)}$ for each
Gaussian center $\mathbf{x}_i$ is then interpolated over its $N$
nearest anchors:
\begin{equation}
    \Delta\mathbf{x}_i^{(0)}
    = \sum_{k\in\mathcal{N}_i} w_{ik}\,\Delta\mathbf{P}_k,
    \qquad
    w_{ik} \propto
    \exp\!\left(-\alpha\,\|\mathbf{x}_i - \mathbf{P}_k^{t-1}\|_2\right)
\end{equation}

\subsection{Joint Online Optimization}
\label{sec:joint_opt}
Given the deformation initialization from Sec.~\ref{sec:anchor_deform_init} and the pose initialization from Sec.~\ref{sec:pose_init_update}, we jointly refine the 3D scene and camera pose. The scene loss is:
\begin{equation}
    \mathcal{L}_{\mathrm{scene}}
    = w_{\mathrm{photo}}\mathcal{L}_{\mathrm{photo}}
    + w_{\mathrm{geo}}\mathcal{L}_{\mathrm{geo}}
    + w_{\mathrm{def}}\mathcal{L}_{\mathrm{def}}
\end{equation}
where $\mathcal{L}_{\mathrm{photo}}$ and $\mathcal{L}_{\mathrm{geo}}$
are masked photometric and geometric rendering losses, and
$\mathcal{L}_{\mathrm{def}}$ regularizes the deformation field, as in~\cite{hayoz2024online}. 
The photometric and geometric losses are defined as the average distance between the rendered $(\hat{I}, \hat{D})$ and input $(I, D)$ color and depth maps over valid pixels $\Omega_t$ and the deformation loss is defined by
$\mathcal{L}_{\mathrm{def}} = \sum_{k} \lambda_k \mathcal{L}_k$,
incorporating local rigidity, isometry, and rotation consistency losses as in~\cite{hayoz2024online}. When the camera is static ($g_t=0$), only
$\mathcal{L}_{\mathrm{scene}}$ is minimized. If $g_t=1$, we additionally optimize an incremental pose update
$\delta\boldsymbol{\xi}\in SE(3)$ via:
\begin{equation}
    \mathcal{L}_{\mathrm{joint}}
    = \mathcal{L}_{\mathrm{scene}}
    + w_{\mathrm{trk}}\,\mathcal{L}_{\mathrm{trk}}
    + w_{\mathrm{reg}}\|\delta\boldsymbol{\xi}\|^2,
\end{equation}
where the $\mathcal{L}_{\mathrm{trk}}$ loss enforces 2D reprojection consistency of the
tracked anchors under the estimated camera motion:
\begin{equation}
    \mathcal{L}_{\mathrm{trk}}
    = \frac{1}{|\mathcal{K}_t|}
      \sum_{k\in\mathcal{K}_t}
      \left\|
          \mathbf{q}_k^{t}
          - \mathrm{proj}\!\left(\mathbf{T}_t\,\mathbf{P}_k^{t-1}\right)
      \right\|_1
\end{equation}

\begin{figure}[t]
  \centering
  \includegraphics[width=0.85\textwidth]{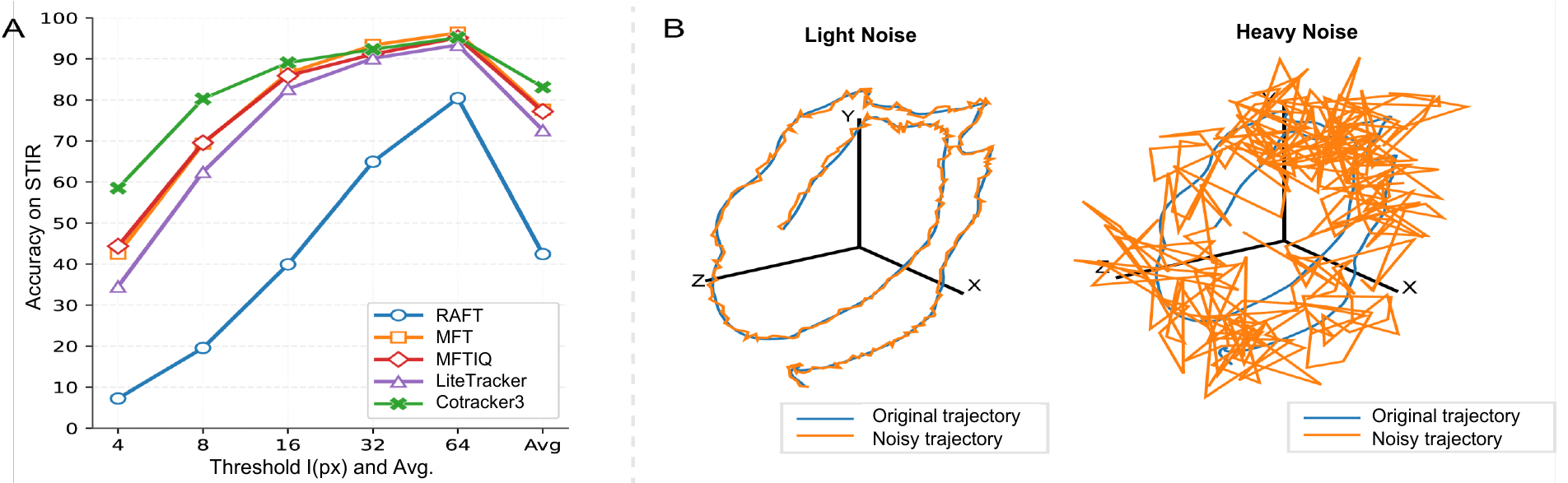} 
  \caption{Illustrating method and experimental design choices. A) 2D tracker selection showing \ac{sota} tracking accuracy across pixel thresholds and average on STIR~\cite{schmidt2025point}, attributed to~\cite{karaev2025cotracker3}. B) Pose noise perturbation strategy applied to sequence P2\_1 from~\cite{hayoz2023learning}. 
Light noise: $\sigma_t=6\times10^{-4}$, $\sigma_r=0.6^\circ$. 
Heavy noise: $\sigma_t=6\times10^{-3}$, $\sigma_r=0.6^\circ$.}
  \label{fig:acc_cotracking}
\end{figure}

\begin{table*}[t]
\centering
\caption{3D reconstruction performance on StereoMIS grouped by pose initialization regime. Online-endo-track~\cite{hayoz2024online} is evaluated only when poses are available. Track2Map (Monocular) uses left RGB with Depth Anything V2~\cite{yang2024depth}; Track2Map (Stereo) is the main setting.}
\label{tab:stereomis_unified}
\small
\setlength{\tabcolsep}{4pt}
\begin{tabular}{llccc}
\toprule
Pose Init & Method & PSNR$\uparrow$ & SSIM$\uparrow$ & LPIPS$\downarrow$ \\
\midrule

\multirow{8}{*}{None}
& Endo-2DTAM~\cite{huang2025advancing} & 15.15 & 0.35 & 0.55 \\
& EndoGSLAM-H~\cite{wang2024endogslam} & 16.67 & 0.52 & 0.45 \\
& ESLAM~\cite{johari2023eslam} & 18.70 & 0.54 & 0.57 \\
& EndoSD-SLAM~\cite{wang2025endosd} & 19.56 & 0.65 & 0.31 \\
& EndoFlow-SLAM~\cite{wu2025endoflow} & 21.96 & 0.59 & \textbf{0.27} \\
& Track2Map (Monocular) & 26.70 & 0.70 & 0.31 \\
& Track2Map (Stereo) & \textbf{27.58} & \textbf{0.745} & 0.274 \\
\midrule

\multirow{4}{*}{Noisy}
& Online-endo-track(light noise)~\cite{hayoz2024online} & 27.31 & 0.727 & 0.349 \\
& Online-endo-track(heavy noise)~\cite{hayoz2024online} & 24.78 & 0.661 & 0.452 \\
& Track2Map (Stereo)(light noise) & \textbf{27.68} & \textbf{0.747} & \textbf{0.276} \\
& Track2Map (Stereo)(heavy noise) & \textbf{27.56} & \textbf{0.742} & \textbf{0.302} \\
\midrule

\multirow{2}{*}{Clean}
& Online-endo-track~\cite{hayoz2024online} & 27.58 & 0.744 & 0.235 \\
& Track2Map (Stereo) & \textbf{27.83} & \textbf{0.750} & \textbf{0.201} \\
\bottomrule
\end{tabular}
\end{table*}

\begin{table}[htbp]
\centering
\caption{Pose estimation on StereoMIS. ATE and consecutive-motion RPE are reported.}
\label{tab:pose_stereomis_endo3r_vs_ours}
\small
\setlength{\tabcolsep}{4pt}
\renewcommand{\arraystretch}{0.95}
\begin{tabular}{lccc}
\toprule
\textbf{Method} & \textbf{ATE(m)} $\downarrow$ & \textbf{RPE}$_r$ $ (deg)\downarrow$ & \textbf{RPE}$_t$(m) $\downarrow$ \\
\midrule
EndoGSLAM-H~\cite{wang2024endogslam} & 0.1077 & 2.4840 & 0.0131 \\
Endo3R~\cite{guo2025endo3r} & 0.0897 & 3.4796 & 0.0354 \\
Track2Map & \textbf{0.0285} & \textbf{0.2903} & \textbf{0.0061} \\
\bottomrule
\end{tabular}
\end{table}

\begin{figure*}[t]
  \centering
  \includegraphics[width=0.9\textwidth]{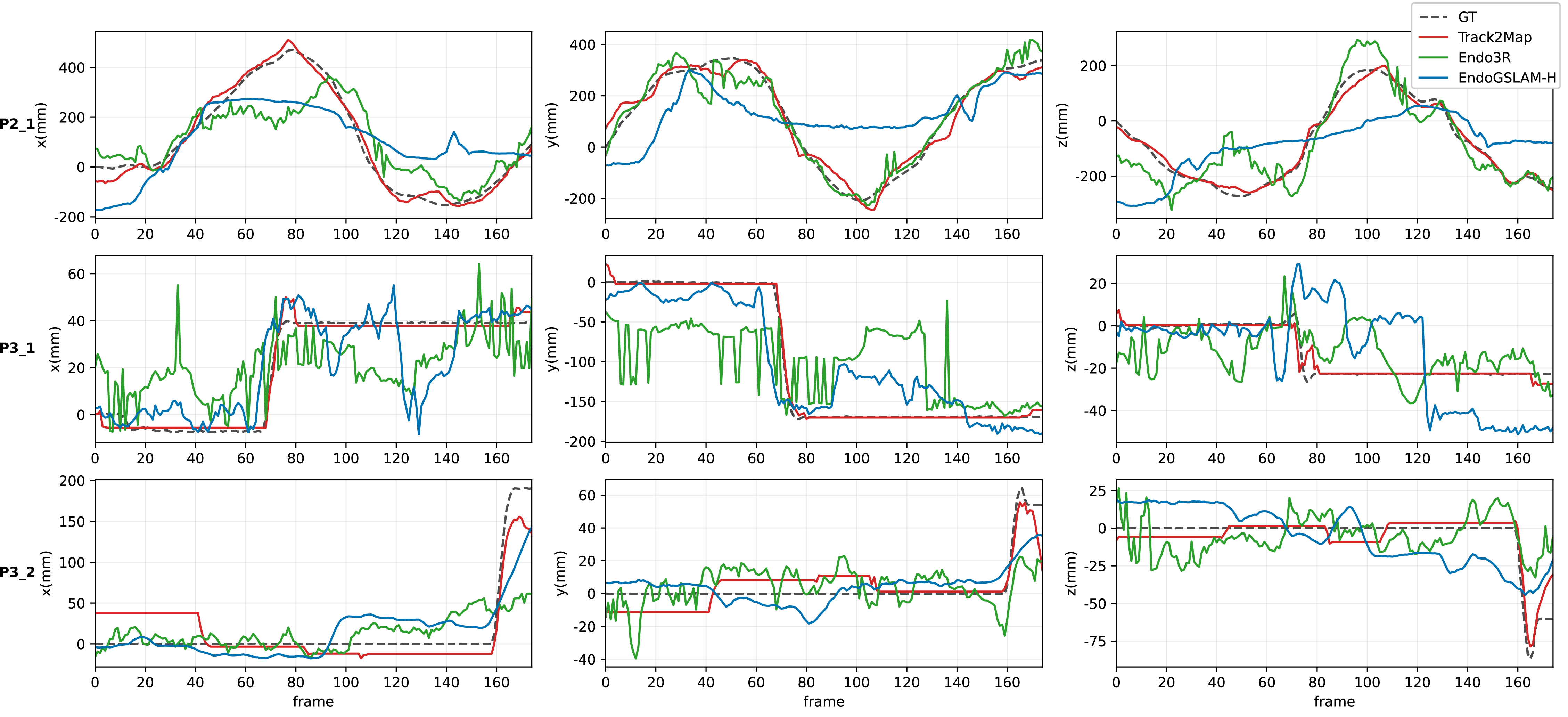}
  \caption{Comparison of estimation of pose on three StereoMIS~\cite{hayoz2023learning} sequences compared to a direct pose estimation network~\cite{guo2025endo3r} and SLAM method~\cite{wang2024endogslam}.}
  \label{fig:pose_estimation}
\end{figure*}

\begin{figure}[t]
  \centering
  \includegraphics[width=0.85\textwidth]{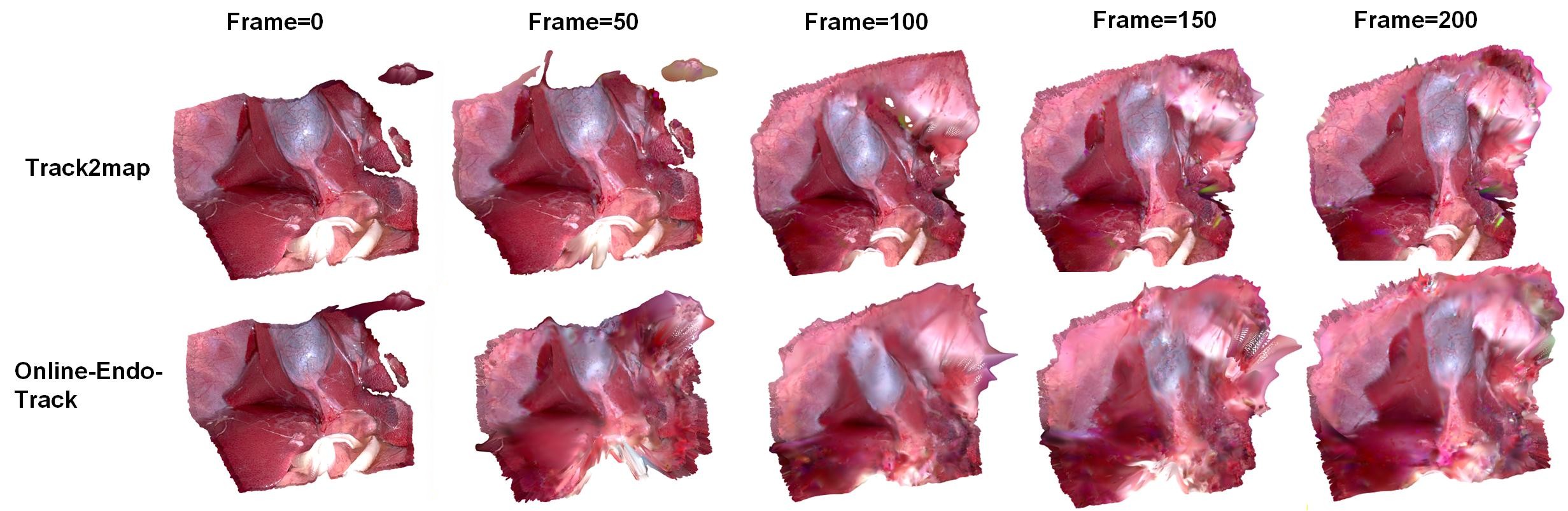}
  \caption{Comparison of reconstruction under heavy noise poses.}
  \label{fig:3dvis}
\end{figure}

\begin{table}[t]
\centering
\caption{Reconstruction ablations on StereoMIS under noisy pose priors.}
\label{tab:pose_ablation_recon}
\small
\begin{tabular}{lcccccc}
\toprule
& \multicolumn{3}{c}{\textbf{Heavy noise}} & \multicolumn{3}{c}{\textbf{Light noise}} \\
\cmidrule(lr){2-4} \cmidrule(lr){5-7}
Method Variant 
& PSNR$\uparrow$ & SSIM$\uparrow$ & LPIPS$\downarrow$
& PSNR$\uparrow$ & SSIM$\uparrow$ & LPIPS$\downarrow$ \\
\midrule

w/o Pose Opt.      
& 25.42 & 0.676 & 0.359
& 27.54 & 0.744 & 0.244 \\

w/o Gating         
& 24.41 & 0.632 & 0.410
& 26.43 & 0.675 & 0.263 \\

w/o Deformation    
& 20.64 & 0.602 & 0.451
& 23.25 & 0.638 & 0.330 \\

\midrule
\textbf{Full Track2Map} 
& \textbf{27.56} & \textbf{0.742} & \textbf{0.302}
& \textbf{27.68} & \textbf{0.747} & \textbf{0.276} \\
\bottomrule
\end{tabular}
\end{table}

\section{Experimental Setup and Results}
\subsection{Experimental setup}
\label{sec:Experiments}
We benchmark 2D point tracker accuracy on STIRC2024~\cite{schmidt2024tracking} to select CoTracker3 as the correspondence module used by Track2Map (Fig.~\ref{fig:acc_cotracking}A). In Tab. ~\ref{tab:stereomis_unified} we evaluate the scene reconstruction quality of Track2Map. Without pose priors, we compare it against SLAM baselines, and with pose priors (clean and noisy) we compare it against \cite{hayoz2024online}. The noise magnitudes injected to pose priors in this experiment can be visualised in Fig.~\ref{fig:acc_cotracking}B. In Tab. \ref{tab:pose_stereomis_endo3r_vs_ours} and Fig.~\ref{fig:pose_estimation} we evaluate camera pose estimation against a direct pose regression model~\cite{guo2025endo3r} and a SLAM baseline~\cite{wang2024endogslam}. We perform an ablation in Tab. ~\ref{tab:pose_ablation_recon} to quantify the effect of pose optimisation, static camera gating, and deformation in our method. 

We use public CoTracker3 online weights~\cite{karaev2025cotracker3} with an 8-frame sliding window, stereo depth from~\cite{wen2025foundationstereo}, a conservative motion-gating threshold of 0.38 (Sec.~\ref{sec:pose_init_update}), and PyTorch on an NVIDIA Tesla V100. Key loss weights are set as: $w_{\mathrm{depth}}{=}10$, $w_{\mathrm{color}}{=}5$, $w_{\mathrm{pose\_prior}}{=}0.25$.


\subsection{Results \& Discussion}
\label{sec:discussion}

Tab.~\ref{tab:stereomis_unified} summarizes reconstruction on StereoMIS under absent, noisy, and clean pose initialization. In the no-pose regime, Track2Map substantially improves PSNR and SSIM over prior endoscopic SLAM and neural mapping baselines, and achieves the strongest overall reconstruction when run in stereo. This improvement is likely driven by two factors: stereo input grounds the reconstruction in a consistent metric scale, and the coupling of online mapping with motion-aware pose updates prevents pose drift during camera-static intervals, stabilizing the incremental mapping process under deformable motion. These components allow the system to recover coherent geometry without relying on robot kinematics or precomputed trajectories. When noisy pose priors are provided, Track2Map has nearly the same reconstruction results as compared to the clean setting. In contrast, methods that depend more directly on the accuracy of pose initialization, such as ~\cite{hayoz2024online}, exhibit larger performance drops under heavy corruption. The qualitative comparison in Fig.~\ref{fig:3dvis} further illustrates this robustness under heavy pose noise: Track2Map maintains more coherent anatomy and fewer visible geometric distortions than the pose-prior baseline, consistent with the quantitative gains in Tab.~\ref{tab:stereomis_unified}. For pose estimation, Track2Map achieves lower ATE and RPE than both Endo3R~\cite{guo2025endo3r} and EndoGSLAM~\cite{wang2024endogslam} (Tab.~\ref{tab:pose_stereomis_endo3r_vs_ours}), with trajectories visually improved in Fig.~\ref{fig:pose_estimation}.

Tab.~\ref{tab:pose_ablation_recon} isolates each component under noisy pose priors. Removing deformation causes the largest drop, confirming that deformation priors are critical for stable reconstruction from limited viewpoints. Disabling motion gating also degrades performance, especially under heavy noise, as local tissue motion can be absorbed into the camera trajectory. Removing pose optimization further reduces quality when pose noise is large. Overall, the full model performs best, showing that deformation modeling, motion-aware gating, and pose refinement are complementary.

\noindent\subsubsection{Limitations and assumptions.} The current implementation of Track2Map is not real time, running at 6 seconds per frame, so we recommend keyframe processing. The motion gate assumes that coherent global flow is more likely to indicate camera motion, whereas heterogeneous or local flow is more likely to indicate tissue/tool deformation. This assumption may fail under optical-axis motion in tubular structures, large coherent tissue motion, tracker failure, substantial tool occlusion, or weakly textured tissue. Finally, evaluation is mainly performed on StereoMIS, so broader generalization should be validated on additional datasets and surgical scenarios.

\section{Conclusion}
\label{sec:conclusion}
We presented Track2Map, an online deformable reconstruction system for surgical video that remains stable when camera pose priors are missing or corrupted. Across StereoMIS, it improves reconstruction quality under clean, noisy, and no-pose settings, and yields more accurate camera trajectories. Robust deformable mapping in RAMIS can therefore be achieved without precise kinematics by leveraging constraints from stereo depth, consistent 2D correspondences, and assumptions about typical endoscopic motion during surgical procedures.

\bibliographystyle{splncs04}
\bibliography{Paper-2834}
\end{document}